\documentclass[sigconf]{acmart}
\usepackage{booktabs} % For formal tables
\usepackage{natbib}
\usepackage{hyperref}
\usepackage{makecell}
\usepackage{hhline}
\usepackage{soul}

\renewcommand\footnotetextcopyrightpermission[1]{}
\usepackage{amsmath}
\usepackage[linesnumbered,boxed]{algorithm2e}
\setcopyright{rightsretained}
\usepackage{ifthen}
\usepackage{comment}

\newcommand{\squishlist}{
   \begin{list}{$\bullet$}
    { \setlength{\itemsep}{0pt}      \setlength{\parsep}{0pt}
      \setlength{\topsep}{-3pt}       \setlength{\partopsep}{0pt}
      \setlength{\listparindent}{-2pt}
      \setlength{\itemindent}{-5pt}
      \setlength{\leftmargin}{1em} \setlength{\labelwidth}{0em}
      \setlength{\labelsep}{0.5em} } }

\newcommand{\squishend}{
    \end{list}  }
    
\newboolean{publicversion}
\setboolean{publicversion}{false}

\ifthenelse{\boolean{publicversion}}{
  \newcommand{\todo}[1]{}
}{
  \newcommand{\todo}[1]{{\color{red}\sf\bfseries #1}}
}

{
  
}

\begin{document}

\title{Towards Fast and Energy-Efficient Binarized Neural Network Inference on FPGA}

% \titlenote{Produces the permission block, and
%  copyright information}
% \subtitle{Extended Abstract}

% \affil[1]{Department of Computer Science and Engineering}
\author[C.Fu et al.]
       {Cheng Fu$^{1,2,*}$, Shilin Zhu$^2$, Hao Su$^{2}$,
       Ching-En Lee$^1$, Jishen Zhao$^{2,*}$
       \\
       $^1$Iluvatar CoreX, San Jose, USA\\
    %   $^2$NVIDIA Research\\
       $^2$Department of Computer Science and Engineering, University of California San Diego, USA\\
       $*$\{cfu, jzhao\}@ucsd.edu\\ 
       }

\begin{abstract}

% \todo{Flow: (1) background of BNN and its benefit (1 - 2 sentences);
%   problem with BNN inference, and why it is important to adapt between
% input reuse and weight reuse (1 - 2 sentences); (3) our proposed
% solutions: we propose a software and hardware co-design of BNN acceleration, which exploits input and weight reuse xxx. We propose an algorithm for weight re-ordering xxx (I assume this is the software design part). Furthermore, we propose an FPGA accelerator architecture that xxx  (1 - 3 sentences); (4) key experiment results (1 sentence).}

Binarized Neural Network (BNN) removes bitwidth redundancy in classical CNN by using a single bit (-1/+1) for network parameters and intermediate representations, which has greatly reduced the off-chip data transfer and storage overhead. However, a large amount of computation redundancy still exists in BNN inference. By analyzing local properties of images and the learned BNN kernel weights, we observe an average of $\sim$78\% input similarity and $\sim$59\% weight similarity among weight kernels, measured by our proposed metric in common network architectures. Thus there does exist redundancy that can be exploited to further reduce the amount of on-chip computations. 

Motivated by the observation, in this paper, we proposed two types of fast and energy-efficient architectures for BNN inference. We also provide analysis and insights to pick the better strategy of these two for different datasets and network models. By reusing the results from previous computation, much cycles for data buffer access and computations can be skipped. By experiments, we demonstrate that 80\% of the computation and 40\% of the buffer access can be skipped by exploiting BNN similarity. Thus, our design can achieve 17\% reduction in total power consumption, 54\% reduction in on-chip power consumption and 2.4$\times$ maximum speedup, compared to the baseline without applying our reuse technique. Our design also shows 1.9$\times$ more area-efficiency compared to state-of-the-art BNN inference design. We believe our deployment of BNN on FPGA leads to a promising future of running deep learning models on mobile devices. 

% Because BNN only contains two values (-1,+1), thus many computations can be skipped and reduced. Traditionally, people used either input reuse or weight reuse, but which one is better depends on the input properties and network architectures.
\end{abstract}




\maketitle

\section{Introduction}

The thriving of Deep Neural Networks (DNN), especially Convolutional Neural Network (CNN), is empowered by the advance of hardware accelerators, such as GPU \cite{CUDA}, TPU \cite{TPU}, and neural network accelerators integrated into various embedded processors~\cite{Sim:2016:TDC}. % In particular, the convolutional neural network (CNN) -- as an important branch of DNN -- can achieve high precision in object recognition and are widely used in various real-time applications, such as computer vision~\cite{Krizhevsky:2017:ICD:3098997.3065386}, speech recognition~\cite{6857341}, and natural language processing~\cite{DBLP:journals/corr/YinSXZ15}.
%autonomous driving~\cite{} and smart home~\cite{}. 
The major challenges of accelerating classical CNNs, which are based on floating-point arithmetic, are (1) off-chip Dynamic Random-Access Memory (DRAM) access power overhead, and (2) on-chip data storage constraints. Many prior works have been proposed to accelerate CNNs by exploiting sparsity \cite{Han} or leveraging data reuse \cite{Eyeriss}. 

Among all kinds of solutions to the above challenges, quantization with reduced bitwidth in network parameters and input data is one of the most promising approaches. Algorithms such as \cite{ZhuHMD16, ZhouYGXC17, Binarynet} have successfully reduced the bitwidth of network weights while maintaining a high precision for image classification tasks. In particular, Binary Neural Network (BNN), a binary quantized version of CNN, has been studied extensively since it can significantly alleviate the DRAM memory access overhead and on-chip storage constraints. In BNN, the multiplication and addition in traditional floating point CNN inference are replaced by more power-efficient, compact, and faster bit operations, which are suitable for reconfigurable logic like Field-Programmable Gate Array (FPGA). However, though the bitwidth in both computation and storage has been considerably reduced, the total number of Multiplication and ACcumulation (MAC) operations still remains the same. For example, binarized VGG-16 neural network ~\cite{DBLP:journals/corr/SimonyanZ14a} has reduced the network storage by around 5$\times$ but it still requires many computations ( $\sim$15.5 Giga MAC operations)
% \todo{Shilin: does this reflect number of bit operations or number of convolution operations? Ans: I have changed the name of operations} 
to do inference on one input image ~\cite{liang2018fp}\cite{umuroglu2017finn}\cite{kim2017kernel}.
% \todo{->Any reference of previous studies or reference to a section that shows our own results}
% The total number of MAC operations 
% \todo{Shilin: this is not true, the number of bit operations is reduced by BNN compared with DNN. You need to define what is arithmetics here, Ans: I mean Multiply-and-add operations here, the total number is not changed} 
% is not reduced compared to floating-point CNN inference although we replaced the floating point multiplication with bitwise operation. 
% In fact, in practice, we usually have to increase the number of parameters for BNN to maintain the classification accuracy as strong as the original floating-point CNN. This added amount of numerical operations would hurt BNN's advantage on power efficiency and latency. 
% Therefore, it is of great importance to reduce the total number of on-chip computations. 
% \todo{rewrite this sentence: an orthogonal direction of the research should be reduce the number of operations of CNN or BNN to further reduce on-chip memory access and computation power.} 

% \todo{Need a transition sentence here.}
% To find an efficient way to reduce the total number of operations for BNN, 
To reduce the number of MAC operations, we leverage the key property of BNN: As the input and kernel weights of BNN are -1/+1, they both exhibit high similarity. Intuitively, the input similarity comes from the spatial continuity of the image to classify, and the kernel similarity comes from the correlation of features represented by different binarized weight kernels \cite{courbariaux2016binarized}. 
% \hao{original: weight kernels can be reordered off-line to achieve high similarity as the reorder of the feature maps is cheap in real-time FPGA. this is not a reason that is intuitive.}. 
% \begin{figure}[bth]
% 	\centering
% 	\includegraphics[width=0.25\textwidth]{fly}
% 	\caption{Percentage of input and kernel similarity across different layers for different applications}
% 	\label{fig:fly}
% 	\vspace{-0.1in}
% \end{figure}
To prove this property of BNN, we studied the similarity of input and kernel across different applications and networks, as shown in Table~\ref{tab:sim}. The kernel similarity is computed based on the re-ordering algorithm described in section ~\ref{reorder}. The average input and kernel similarity ratio is ranging from 78\% $\sim$ 84\% and 59\% $\sim$ 64\% for network models \cite{Binarynet} \cite{DBLP:journals/corr/LinCY13}. However, if the weights of BNN are binarized but the activations are finely quantized (Table~\ref{tab:sim}), which is a favorable setting in many current works~\cite{DBLP:journals/corr/CourbariauxBD15}, %However, it has been shown that activation binarization has larger impact on the performance BNN inference, and thus many existing works consider only binarized weight with quantized fixed-point input activation. 
%In such case as shown in Table~\ref{tab:sim}, 
the kernel similarity is much higher than the input similarity. % due to their difference in quantization level. 
In other words, we see that the degree of these similarities highly depends on the dataset and network architectures.

Based on these observations, we propose an architecture of BNN accelerator that leverages input and kernel similarities to reduce the number of MAC operations at inference time. 
Instead of directly computing the XNOR between the input activation and kernel weights, we first check the input or kernel weight difference between the current and previous computation stage, which focuses on different image regions or different weight kernels, and then reuse the results from the previous computation. % Our computation unit will update the old result based on this difference. 
Thus, the data buffer access or MAC operations can be bypassed if there is no difference from the previous stage. Our analysis shows that 80\% of the computation and 40\% of the buffer access on average can be skipped in this way. 
% \st{if their is no difference between previous execution} \todo{(how much percent (on average of our evaluated NNs) there is no difference between previous execution?)}. 
% Furthermore, by leveraging the input and kernel similarity, the performance of BNN inference can be accelerated by t.t$\times$ and t.t$\times$, respectively in comparison to state-of-the-art.
As a result, our design can reduce the total power consumption by around 17\% and on-chip power consumption by 54 \% in comparison to the one without using our reuse method. Our design is also 1.9$\times$ more area-efficient compared to the state-of-the-art BNN accelerator.  

In addition, we observed that similarites can vary for different applications. Therefore, we provide analysis and insights to pick the better one from our proposed two reuse strategies for different datasets and network models. 

\begin{table*}[t]
  \caption{Input and kernel similarity ratio across different networks and datasets. A=(8,4) means 8-bit fixed point activation input including 4-bit fractional part. LeNet-5 and NIN are trained on XNOR-Net \cite{xnornet}.}
  \label{tab:sim}
  \begin{tabular}{cccccc}
    \toprule
    Dataset & Network & Min Input Sim (\%) & Avg Input Sim (\%) & Max Input Sim(\%) & Kernel Sim (\%) \\
    \midrule 
    MNIST & LeNet-5 & 66.6 & \textbf{79.3} & 88.6 & 59.8\\
    \hline
    MNIST & LeNet-5, A=(8,4) & 10.6 & 37.5 & 67.0 & \textbf{59.8}\\
	\hline
    Cifar-10 & BinaryNet & 59.6 & \textbf{78.6} & 95.6 & 58.8\\
    \hline
    Cifar-10 & BinaryNet, A=(8,4)& 1.8 & 17.3 & 72.2 & \textbf{58.8}\\
	\hline
    Cifar-10 & NIN & 51.3 & \textbf{83.9} & 97.2 & 64.5\\
    \hline
    Cifar-10 & NIN, A=(8,4) & 2.7 &  23.5 & 66.7 & \textbf{64.5}\\

  \bottomrule
\end{tabular}
\end{table*}

To sum up, we make the following contributions: 
%Our main contributions are listed as follows:

% \begin{itemize}
\squishlist
	\item We analyze the input and kernel similarity in BNN across different applications. We also show that the degree of similarity depends on datasets and network architectures and we generate insights to select the best strategy.
    
	\item To the best of our knowledge, we are the first to exploit input and kernel similarity to reduce computation redundancy to accelerate BNN inference. We propose two types of novel, scalable, and energy-efficient BNN accelerator design to leverage different types of similarity. Our comparison between these two accelerators provide guidelines to pick the best one or even combine them together.
    
%     \todo{-- input and kernel similarity}. \todo{Describe what do you mean by "reuse strategy".} \todo{Rewrite this sentence: Each of them can be fit for different application requirement.}
    
	\item Our implementation indicates that by exploiting similarities in BNN, we can push the efficiency and speed of its inference to a higher level.
% 	save on-chip operations by up to 80\% and reduce weight buffer access by 40\% compared to the design without any computation reuse. 
\squishend
% \end{itemize}
The code of this work will be made public online. The rest of this paper is organized as follows: Section 2 gives an introduction of CNN and BNN inference; Section 3 describes our motivation to exploit input and kernel similarity and presents our method; %of exploiting input and kernel similarity and weight re-ordering algorithm BNN accelerator design; 
Section 4 provides the hardware architecture of the accelerator; Section 5 reports our experiments and findings; Section 6 reviews previous work on traditional accelerator design; finally, we discuss potential future work in Section 7 and conclude the paper in Section 8.
    
\section{Background}
\subsection{Convolutional Neural Networks}
Convolutional Neural Networks (CNNs) are commonly used for image processing, object recognition and video classification ~\cite{DBLP:journals/corr/SimonyanZ14a}. A convolutional layer implements a set of kernels to detect features in the input image. A kernel is defined by a set of weights $W$ and a bias term $B$. Each convolutional layer applies multiple kernels on the input where each kernel scans through the input in a sliding way, resulting in multiple output feature maps (\textit{ofmap}). Note that, unlike what happens in Fully-Connected (FC) layers, the weights of a kernel are shared across different locations on the input. Formally, suppose the input vector is $X$ and weight vector of a kernel is $W$, then the \textit{ofmap} $O$ of this convolution operation is the dot product between them, added by the bias $B$, and followed by a non-linear activation function $g(\cdot)$ as shown in Equation ~\ref{eq:1}:
\begin{equation}\label{eq:1}
O = g(W \cdot X + B)
\end{equation} 

In CNNs, a pooling layer is usually added after a convolutional layer. FC layers are often appended after several stacked convolutional blocks. During training, the ground-truth output serves as a supervision signal to learn parameters $W$ and $B$ by minimizing a loss function. After a CNN has been trained, the network inference is applied to the test image. Previous work ~\cite{7827589} shows that the computation of the CNN inference is dominated by the convolution operation, which is our main focus in this work.

Thanks to the idea of weight sharing, lots of computation are actually not necessary because convolution is naturally a sliding-based operation. The goal of this paper is to reduce the computation cost of convolutions by exploiting what we have already computed so that we can reuse them instead of computing repeatedly.

\subsection{Binarized Neural Networks}
Recent studies identify that there is no need to employ full-precision weights and activations since CNN is highly fault-tolerant ~\cite{8416871}; we can preserve the accuracy of a neural network using quantized fixed-point values, which is called quantized neural network (QNN) \cite{DBLP:journals/corr/HubaraCSEB16}. An extreme case of QNN is Binarized Neural Network (BNN) ~\cite{xnornet}, which adopts weights and activations with only two possible values (e.g., -1 and +1).
%The other approach is Binarized Neural Network (BNN), which adopts weights and activations with only two possible values (e.g., -1 and +1). The BNN approaches can be viewed as an extreme example of the quantized CNN models. 
%commonly used for hardware acceleration. 
The most widely used binarization strategy is called deterministic binarization, as illustrated in Equation ~\ref{eq:2}. This strategy is preferred because it is suitable for hardware accelerations. 

\begin{equation}\label{eq:2}
x_{b} = \text{Sign}(x)
\end{equation} 
Here, $x$ can be any weight or activation input and $x_{b}$ is its binarized version. %Back-propagation becomes challenging because the derivative of sign function is ill-posed. 
% %In recent years, people 
% %Recent studies proposed
% %lot of possible 
% %solutions to this issue, such as the straight-through estimator (STE). \cfu{any correlation with our paper? we need to mention the step of doing bnn inference, batch normalizaiton}
It has been shown that input activation binarization causes much more degradation to the accuracy of BNN classification compared with weight binarization \cite{courbariaux2016binarized, zhou2016dorefa, zhu2018binary}. Thus we consider two BNN configurations: (i) Both input and weights are binarized. (ii) Input is quantized to fixed-point values and weights are binarized. These two configurations also affect our design choice and will be described in Sec 3.2.
As for implementation of this paper, we mainly 
%mainly 
focus on accelerating a BNN model
%architecture 
developed by Courbariaux et al. in \cite{Binarynet}. However, our proposed scheme can be employed on any BNN, which has convolution operations during inference phase. Throughout the rest of this paper, we represent the input activation vector $X$ of the BNN as $\text{IA}(h,w,c)$, corresponding to the horizontal index, vertical index, and channel index. We further denote the weight vector $W$ by $W(r,s,c,k)$, corresponding to horizontal index, vertical index, channel index and kernel index.

%in this work. 

% \todo{Nakahara et al.~\cite{Nakahara} presented a modified BNN version of YOLOv2 to perform real-time localization and classification. -> Is this an FPGA acceleration work or not? If not, should be discussed somewhere else. Ans: Yes, its an FPGA'18 paper}

\section{Design Principles}
In this section, we first introduce the objective of our method with a key observation on BNN's property (Sec 3.1), which drives our proposed reuse principle (Sec 3.2). Moreover, we solve an offline optimization problem to further improve the gain (Sec 3.3).
\subsection{Motivation}

% Considering a case when the DRAM can provide unlimited bandwidth, but only a limited number of multipliers are available on-chip; so even though they can achieved 100\% utilization, we still get a fixed GOPS.

To realize a design that can efficiently accelerate BNN inference, typically people tend to optimize an objective called \textit{throughput} which can be described by  \textit{frame per second} (FPS) as Equation ~\ref{eqfps}:

\begin{equation}\label{eqfps}
  \text{FPS} = \frac{ \# \text{Multipliers} \times \text{Utilization}}{ \# \text{Ops\_per\_image}}
\end{equation} 
where \textit{Utilization} indicates the ratio of time for multipliers doing inference over the total runtime. To increase the FPS, previous BNN works seek to increase the number of multipliers by reducing the control overhead. Other works exploit a highly parallelized computation architecture that can increase the \textit{Utilization}. But another orthogonal direction for increasing the FPS is by reducing the number of $Ops\_per\_image$, which has not been fully exploited in current works. In this paper, we notice that a large amount of computation redundancy exists in BNN inference. Thus, our approach aims to reduce the number of $Ops\_per\_image$ -- that is to utilize the input or kernel similarities which will be discussed in section ~\ref{reusestrategy}.
Our work also has the advantage of reducing on-chip power consumption. Specifically, the data buffer access and computation power can be saved as a result of the reduced number of \textit{Ops\_per\_image}.

\subsection{Similarity Inspired Reuse Strategy}
\label{reusestrategy}
%\todo{An overview of key insights and design principles: first explain why we need to propose a new algorithm optimization, i.e., the goal and insights of our software design; then summarize what we have proposed.}
%\hao{people won't understand what kernel similarity means before you present the algorithm. you should introduce the algorithm first before you conclude that there is high similarity among weights. }
%\todo{Shilin: Cut the head into half into two categories.}
Recall that our objective is to reduce the number of \textit{Ops\_per\_image} to maximize the throughput. Unlike floating-point values in CNNs, BNN has binarized weight and input after the model is trained. Thus, BNN has only two values (-1/+1), which means we have 50\% chance of having the same value if we pick random two numbers in weight or input. In what follows, we introduce two types of similarity and show our statistical results.
% To realize this , we exploit the similarity exist in BNN as 

\noindent\textbf{Input similarity: } Input similarity naturally exists in various datasets when we use BNN to classify images. Most natural images have spatial continuity and adjacent pixels are similar. In consequence, the binarized \textit{ofmaps} are also very likely to be spatially similar.

\noindent\textbf{Kernel similarity: } kernel similarity comes from the affinity of features represented by different binarized weight kernels. It has been shown in literature that weight kernels of BNN are highly similar (only 42$\%$ are unique on Cifar-10 in \cite{courbariaux2016binarized}). The kernel similarity can be further optimized by using the algorithm introduced in Section \ref{reorder} which can be computed off-line. 

% Yet, the default order of convolution does not necessarily provide sufficient kernel similarity. In order to enhance the kernel similarity in BNNs, we propose a weight re-ordering algorithm to maximize kernel similarity by permuting the weight along the filter dimension ($k$). In essence, our permutation process can be considered as a graph optimization problem and performed off-line as described in section ~\ref{reorder}.

Based on these properties, we did an experiment on multiple BNN models to evaluate the input and kernel similarities which are defined as Equation ~\ref{eqia} and ~\ref{eqw}.

\begin{equation}\label{eqia}
  \text{IA}(h,w,c) = \text{IA}(h,w-1,c), 0 < w \leq W_m
\end{equation} 
\begin{equation}\label{eqw}
% \begin{split}
  W(r,s,c,k) = W(r,s,c,k-1), 0 < k \leq K
% \end{split}  
% \end{center}
\end{equation}
The input or weight \textit{similarity ratio} is defined as the input or weight values that are subjected to Equation \ref{eqia} and Equation \ref{eqw} over the size of total input and weight values. The results are illustrated in Table ~\ref{tab:sim}. The reported kernel similarity is optimized by using the algorithm as described in section ~\ref{reorder}.

As we can see in Table ~\ref{tab:sim}, the BNN has different average similarity ratio in both weight and input across different models. For BinaryNet \cite{Binarynet}, the average input and kernel similarity are $\sim$78\% and $\sim$58\% respectively, which indicates a high computational redundancy in BNN inference and so does NIN on XNOR-Net \cite{xnornet}. For BNN trained by XNOR-Net on MNIST, the average similarity ratio on input and weights are $\sim$79.3\% and $\sim$59.8\% respectively. We also observe that many of the BNN models are implemented with fixed-point input activations to maintain high classification accuracy \cite{zhou2016dorefa}. For BNN with fixed-point input and binarized weight, the input similarity is much lower than kernel similarity as shown in Table ~\ref{tab:sim}. Therefore, which one is better varies case by case. The comparison and determination can be done off-line after the network and dataset are obtained. 

Depends on the insights generated by our experiment results, we develop two types of neural networks acceleration strategies which exploits either weight or input similarity to reduce redundant computations. We now introduce two computation reuse strategies to leverage these two types of similarity respectively.

\begin{figure}[bth]
	\centering
	\includegraphics[width=0.5\textwidth]{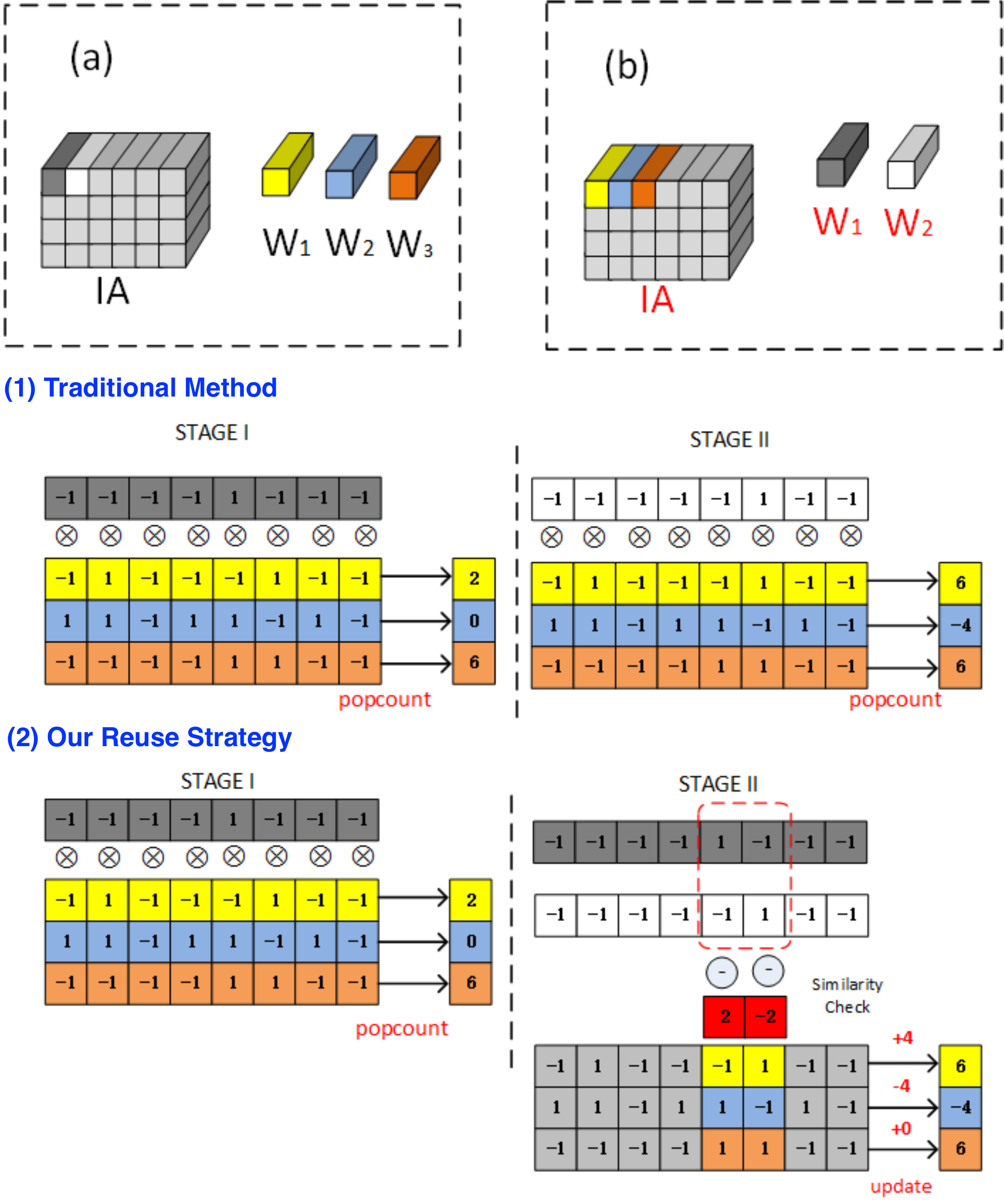}
	\caption{Method of exploiting input and kernel similarity. (1) shows the computation without using the reuse strategy. (2) illustrates the computation with reuse strategy. The computation of STAGE II reuse the computation result from STAGE I by exploiting (a) input similarity (b) kernel similarity}
	\label{fig:Method}
	\vspace{-0.1in}
\end{figure}

In Figure ~\ref{fig:Method}, we use a simple example to illustrate the idea of computation reuse. CNN inference involves multiple dot products between input and kernels and they are computed in a sequential manner. The two stages shown in Figure ~\ref{fig:Method} represent two consecutive dot products during convolution. As shown in Figure ~\ref{fig:Method} (1), the traditional way of computing dot product between the input pixels and different $1\times1$ weight kernels is by doing bitwise XNOR and then popcounts the resulted vector for accumulation for both \textit{STAGE I} and \textit{STAGE II}. With the computation reuse strategy (Figure ~\ref{fig:Method} (2)), we still compute the result in the traditional fashion in \textit{STAGE I}. In \textit{STAGE II}, instead of computing the dot product again in the traditional way, we can save computations by updating the result from \textit{STAGE I}. This method can be applied to leverage either input or kernel similarity. Note that here we use $1\times1$ kernel for intuitive illustration, but our reuse strategy can be easily extended to larger weight kernels using parallelism, such as $3\times3$ kernels used in our experiments (Section ~\ref{eva}).

The details of the two methods are further analyzed below.

\vspace{3pt}
\noindent\textbf{Input Reuse:} For input reuse, assume the computation of \textit{STAGE I} is finished in Figure ~\ref{fig:Method}(2). For the next dot product computation in \textit{STAGE II}, we can first check the difference between the current input (\textit{STAGE II}) and the previous one (\textit{STAGE I}). Then, we can update the previous result from \textit{STAGE I} based on this difference. In this way, the number of required bitwise operations can be greatly reduced compared to the traditional way of computation if the inputs exhibit high similarity. The computation of similarity needs to be done only once, compared with directly computing dot product repeatedly. Besides, only a small subset of weight (colored ones shown in \textit{STAGE II} of Figure ~\ref{fig:Method}(2)) needs to be read out from on-chip memory. 
% However, the extra overhead of this method is the conditional popcounts which means the update direction depends on the sign of the binarized weight.
% once the segment of input is the same, not only the computation but also the weight bank access can be bypassed. As is shown in \textit{STAGE II} of Figure ~\ref{fig:Method}(a), only the colored weight needs to be read out from the on-chip memory. 

% while each weight channel needs to be assigned with a reuse register to save the computation based on the result from previous execution. 
% \cfu{do we need this ? }

% However, the extra overhead is the conditional popcounts which means the update direction depends on the sign of the binarized weight.

% Computing the similarity and judging whether two numbers are the same consumes less power and resource as it needs to be done only once between the two pixels instead of doing XNOR between the input and all the weight chunks $W_k$ in Fig.1 (a). However, the extra overhead is the conditional popcounts which means the update direction depends on the sign of the binarized weight.

\vspace{3pt}
\noindent\textbf{Weight Reuse:} Different weight values at the same position of different kernels, i.e., weights at $(r,s,c,k_i)$ and $(r,s,c,k_j)$ ($k_i$ and $k_j$ denote the indices of two different kernels), exhibits high similarity. This similarity across different weight kernels can also be exploited in the similar way as the input reuse strategy. 
For the weight reuse strategy, the process of reuse shares the same principle with the input reuse. As shown in Figure ~\ref{fig:Method} (b), instead of computing the difference between input activations, we first check the difference between kernels and then update the previous dot product result accordingly.

%The computation of weight reuse strategy is symmetric to the input reuse strategy. The consecutive execution of the dot product is between the same input and different weight kernels. As is shown in Figure ~\ref{fig:Method}(b), instead of compute the difference between input activation, we first check the difference between weights and updates the result from previous execution.

Moreover, we find that the original computation order of the dot product between input and kernels can be further optimized off-line to achieve high degree of kernel similarity ratio, which will be discussed below.

\subsection{Improve Weight Reuse by Optimization}
%\hao{the writing of this subsection is roughly acceptable. }
\label{reorder}
\begin{figure}[bth]
	\centering
	\includegraphics[width=0.4\textwidth]{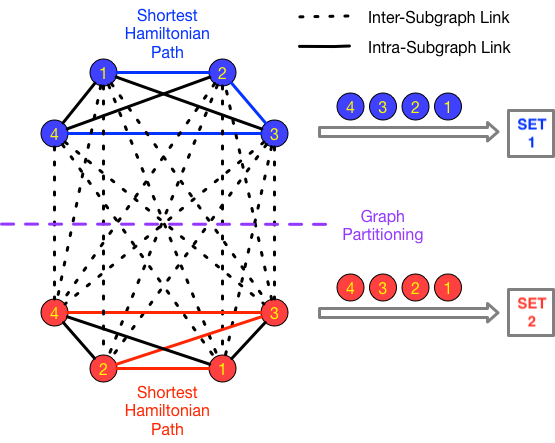}
	\caption{Graph optimization illustration to optimize convolution order of kernels.}
	\label{fig:graph_opt}
	\vspace{-0.1in}
\end{figure}

\begin{algorithm}
 \textbf{Given}: A set of convolution kernels $V$ within a layer, and $K$ partitions required\\
 %PEs available in FPGA\\
 \textbf{Output}: $K$ sets of optimized order of kernels where each set has size $|V|/K$\\
 Construct a graph $G(V,E,W)$ where $w(v_{i},v_{j})=\#\text{Params}^{\text{Diff}}_{i,j} / \#\text{Params}_{i,j}$ (dissimilarity)\;
 Partition $G$ into $K$ subgraphs by maximizing $\sum_{e_{ij} \in \text{inter-subgraph-links}} w(v_{i},v_{j})$, each denoted as $G_{k}, k=1, ..., K$\;
 \For{each $G_{k}$}{
  Find the shortest Hamiltonian Path $H_{k}$ in $G_{k}$\;
 }
 \textbf{Return: }$K$ sets of optimized kernel order as $\left \{H_{k}, k=1, 2, ..., K \right \}$\;
 \caption{Weight Reuse Optimization}
\end{algorithm}

Although regular weight reuse can accelerate BNN inference reasonably, we find that it is possible to optimize the weight reuse by re-ordering the convolution on kernels in each layer. In other words, the default computation order of kernels may not be good enough. Here we develop an algorithm to find a better order of convolutions using graph optimization. As shown in Fig. \ref{fig:graph_opt}, we build a graph $G(V, E, W)$ where each vertex $v \in V$ corresponds to one kernel. Two vertices are connected by link $e \in E$ with weight $w \in W$ where $w$ represents degree of dissimilarity between two kernels. To find the optimal order of convolutions, we need to search for the scheduling where the total dissimilarity is minimized, i.e., total similarity is maximized.

\subsubsection{Graph Partitioning. } We first partition the graph into several subgraphs. The reason is that our proposed architecture has limitations on the number of kernels for reverting in each computation unit (i.e., Set 1 and 2 in Fig. \ref{fig:graph_opt}). Suppose there we partitioned all the kernels into $K$ subset, then each computation unit will work on $|V|/K$ kernels. To partition the original graph into $K$ subgraphs, we maximize the summed weight of links in between subgraphs. In other words, we maximize the dissimilarity in between group of kernels so that similarity is maximized within each group.

\subsubsection{Sub-Graph Optimization. } For each subgraph containing $|V|/K$ vertices, we compute the shortest Hamiltonian path thus the accumulated dissimilarity is minimized along the path. Here we use a greedy approach to solve Hamiltonian path problem in an efficient way because the graph can be very large if there exists a lot of kernels within a layer (e.g., 1024), although non-greedy approach may be viable when the graph size is small, since this is optimized offline after BNN has been trained. The complete algorithm is shown in Algorithm 1.

After the above optimization, we can get the optimized order of convolution in terms of kernel indices for each computation unit. Then we will send this indexing map to the hardware in order to process it. In this design, in order to alleviate the hardware complexity and the overhead of the \textit{ofmaps} reverting process, we put a limitation on $|V|/K$ to be 64.  More complicated design to optimize partitions may be considered in future work. 

\section{Hardware Architecture}
% \hao{this section is not too bad. howver, you have many long sentences with more than 25 words. these sentences are generally broken or have minor grammar mistakes. break them into shorter ones.}

% \todo{It is nice to have a leading sentence summarizing what we propose as below. But the same as Sec.3, we need to explain WHY first. Maybe also add one more sentence about WHAT. Always start from high-level overview of goals and design principles, before getting into details. That will make our insights and novelty stand out. Don't expect readers to dig them out in the description of details.}

In this section, we introduce a hardware architecture that exploits the BNN similarity to save computations and memory accesses. Recall that we have two types of reuse strategies discussed in the above section and so we present two types of accelerators that leverage input and kernel similarity, respectively. 
% The ways of exploiting similarities for these two accelerators are very similar.

\subsection{Input Reuse Accelerator}
% \hao{this sec roughly ok.}
% The hardware architecture of our BNN accelerator is shown in Figure 

\begin{figure}[bth]
	\centering
	\includegraphics[width=0.5\textwidth]{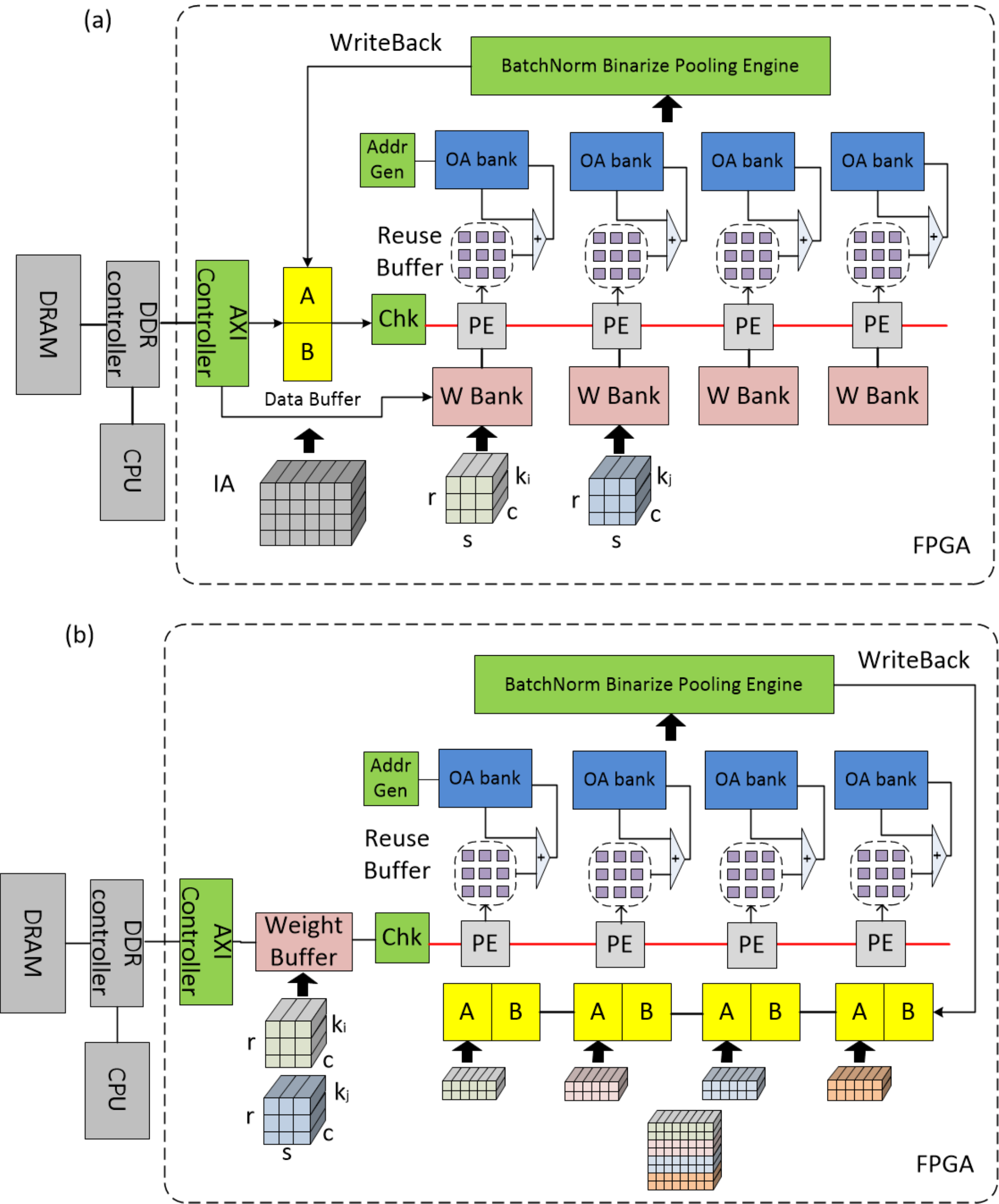}
    \caption{Block diagram of (a) input reuse accelerator and (b) weight reuse accelerator}
	\label{fig:block}
	\vspace{-0.1in}
\end{figure}

% \begin{figure}[bth]
% 	\centering
% 	\includegraphics[width=0.5\textwidth]{figure/fig3.png}
% 	\caption{Block diagram of the input and weight reuse accelerator}
% 	\label{fig:fly}
% 	\vspace{-0.1in}
% \end{figure}
The block diagram of the architecture that exploits the input similarity accelerator is shown in Figure \ref{fig:block} (a). For the input reuse accelerator, the execution has mainly three stages -- data loading, computation, and accumulation. For the data loading stage, the input and weight data will be read from off-chip memory and be stored into the data buffer and weight memory banks (WBank). During the execution of a binarized convolution layer, the input data will be read from one data buffer and written to another equally size buffer. 
% This is originally the idea of layer-fusion technique proposed in \cite{7783725} for fixed-point CNN that can be easily realized by BNN accelerator due to the lower storage requirement \hao{rewrite}. 
% \todo{Shilin: this sentence is too long.} 
The read or write mode of data buffer A and B (Figure ~\ref{fig:block}) will be switched for the computation of different layers so the input of each layer does not need to transfer back and forth between on-chip and off-chip memory.
% The data will be ping-pong back and forth between the Data Buffer A and B for the computation of different layers. 

During the computation stage, the entire system works in a producer and consumer fashion. The producer is the checking Engine (Chk) which is implemented as a bitwise C-by-C subtraction logic for checking the current input versus the previous one. For the computation of the first input during computation stage, which is corresponding to the \textit{STAGE I} discussed in Section ~\ref{reusestrategy}, Chk will broadcast the original input value and the Processing Elements (PE) will compute the result in the traditional way by using XNOR and popcount. For the rest of the input, we will use the reuse method for the computation, which is consistent with \textit{STAGE II} mentioned in Section ~\ref{reusestrategy}. During the reuse computation, Chk subtracts  the current input with the previous one to check the difference. Once the checking is failed, the Chk will broadcast the subtraction result to all the PEs through a broadcasting bus. PE will read the weight out of the Wbank and scan the bus to find the different elements and update the reuse buffer which contains the result of last execution. 

The different input values will be executed by $n$ different PEs simultaneously. Each PE is assigned with a reuse buffer, a Wbank, and an Output Activation bank (OAbank). The storage of weight is partitioned in kernel or $k$ dimensions,
% $(r,s,c,k_i)$ in different $k$ dimensions \todo{Shilin: different k dimensions is not a correct notation.} is stored in different Wbank 
so that the \textit{ofmaps} result will not interleave across different OAbanks. Once the current pixel has finished broadcasting, the accumulation stage will begin. The address generator and accumulator will collect the results in the reuse buffer and accumulate them into the corresponding position of OAbank.

\subsubsection{Address Generator and Accumulator}

The address generator calculates the destination address for different intermediate result in the reuse buffer. The OAbank accumulation controller will collect the result in reuse buffer and reduce them into the correct positions in OAbank which indicates by the address generator. The address of the \textit{ofmap} $(h_o,w_o,c_o)$ (subscript $o$ denotes output) of the given input locating at $(h,w,c)$ and weight locating at $(r,s,c,k)$ can be calculated as $(h-r,w-s,k)$ or $(h-r+1,w-s+1,k)$ if padding mode is enabled.
% \todo{Shilin: check this. 'c' is index, not entire channel. Make sure you are talking about index! Otherwise you need to define new terms}. This reduction method allows the design to handle weight kernels with arbitrary size. 
% \todo{Shilin: where is the channel and batch index? If you remove them, we need to state this change first.}

\subsubsection{Batch Normalization Engine}

Once the computation of the current layer is finished, the batch normalization engine will concatenate the \textit{ofmaps} results from different OAbanks and normalize the output by subtracting the normalization factor before binarizing the value into -1/+1. Our strategy of doing batch normalization and pooling is similar to previous BNN acceleration work like \cite{umuroglu2017finn}\cite{zhao2017accelerating}. Batch-normalization and activation functions are done together by comparing to the normalization factors across different \textit{ofmaps} computed offline and then the pooling is done by using lightweight boolean AND operator. The entire batch normalization engine in our design for input similarity accelerator consumes 1541LUT and 432FF for a PE size of 8. 
% \todo{Shilin: double check it. Did you define LUT and FF previously? Ans: this is correct}

\subsection{Weight Reuse Accelerator}

For the weight reuse accelerator, the architecture is very similar to the input reuse accelerator. First, as is shown in Figure \ref{fig:block} (b), different lines of the input activation (IA) will be evenly distributed across PEs instead of different weight kernels for input reuse accelerators. But still, two equally-sized buffers will be assigned to each PE. The IA data will be read from and written into two separate buffers as the input reuse accelerator. 
% \todo{Shilin: better to remove this A, B notation.} 
Second, instead of broadcasting the input difference to PE like the input reuse accelerators, weight reuse accelerator broadcasts the difference between the weight kernels to PE. 
% \todo{Shilin: 'between weight channel' is a wrong statement.}

We first pre-process the weights off-line by using the algorithm introduced in Section \ref{reorder} to reorder the weight kernels with different $k$ dimensions to produce similarity. The hardware allows the weight kernel to be executed out-of-order in a given re-ordering range 
% (\todo{Shilin: explain this range}) 
as we will revert the sequence of the \textit{ofmaps} on-chip. Larger reordering range can achieve even higher degrees of similarity among weight kernels but also introduces higher \textit{ofmap} reverting overhead. The sequence information of the permuted weight kernels needs to be loaded on-chip for reverting the \textit{ofmaps}. But overall, assuming the reordering range is 64, the sequence information overhead is 10K bits which is less than 0.2\% of the size of the weight kernels and thus can be ignored. 

We also replace the representation of -1/+1 in weight kernel with "same" (0) and "different" (1), except for the first dot product computation. In this way, the on-chip checking process can bypass the subtraction logic.
% \hao{too long, break into a few sentences}.
Once the computation begins, the weights will be loaded into the weight buffer and the input activation will be distributed to the IA buffers on-chip. The checking engine does not need to do subtraction as the weights have been pre-processed off-line in the representation of in "different" (1) versus "same" (0). The first weight kernel to be computed will still use the original value with XNOR and popcount for accumulation.
% \todo{Shilin: I suggest to change 'weight channel' to something more intuitive.} 
It will also be stored in the checking engine as a \textit{weight base} which will be continuously updated during the checking process. The goal of the \textit{weight base} is to keep the latest version of the real weight value to recover the weight difference during the computation.
% \todo{Shilin: we may need to explain one more sentence on this weight base.} 
For the rest of the computations, we begin to use the weight reuse strategy for computation. The Chk will scan the weight value to check the similarity. Once the similarity check fails in the computation, the Chk will generate the weight difference based on the \textit{weight base} vector and broadcast the weight difference to all the PEs where different lines of the input activation are stored. In the meanwhile, the Chk will update the \textit{weight base} which is the real weight data used for the following computation.

The address generator is still used for the calculation of a uniform address for PE reduction once a weight kernel finishes broadcasting.
% \todo{Shilin: I suggest to change 'weight channel' to something more intuitive.} 
As the input is not duplicated in different IA buffer, some results in OAbank are partial sums which need to be further reduced in the last stage before the batch normalization and pooling. The final batch normalization engine will finish the last reduction before the batch normalization and \textit{ofmap} reverting process. The final result will be stored back into the data buffer once the execution is finished. Overall, the design of weight reuse acceleration is a symmetric version of input reuse accelerator where we exploit kernel similarity across different. The differences between input reuse and weight reuse accelerator are mainly in the reduction and reverting logic. More complicated design can achieve better parallelization which is left for future work.

% \todo{Shilin: needs to make high-level summary.}
% The batch normalization and pooling engine now serves the needs to do the final reduction for the partial summation in each OAbank before revert back the sequence of the ofmaps. 

% The hardware also allows the weight kernel to be executed out-of-order in a given re-ordering range 
% % (\todo{Shilin: explain this range}) 
% as we will revert the sequence of the \textit{ofmaps} on-chip. Larger re-ordering range can achieve even higher degrees of similarity among weight kernels but also introduces higher \textit{ofmap} reverting overhead. The sequence information of the permuted weight kernels needs to be loaded on-chip for reverting the ofmaps. But overall, assuming the re-ordering range is 64, the sequence information overhead is 10K bits which is less than 0.2\% of the size of the weight kernels and thus can be ignored.

\section{Evaluation}
\label{eva}
%\hao{roughly acceptable. many grammar mistakes. needs polish.}
In this section, we evaluate our proposed reuse strategies on a real FPGA board. The following results show increased performance by using our method compared with benchmark BNN model in terms of speed and energy efficiency.
\subsection{Prototype Implementation}
To evaluate our design, we implement two types of BNN inference accelerators which exploits input and weight reuse strategies respectively for convolutional layers (conv). The process of decision-making between input or weight reuse strategy depends on the performance of the two types of accelerators on a given model. We test the design on the BinaryNet -- an inference model for CIFAR-10 \cite{Krizhevsky09} ($32 \times 32 $ color images) classification. The pre-trained BinaryNet \cite{Binarynet} neural network is from the open-source code of \cite{zhao2017accelerating}. The summary of the workload is listed in Table ~\ref{tab:workload}. The BinaryNet for CIFAR-10 model can achieve $11.19\%$ testing error. Our design can also be used to accelerate BNN models other than BinaryNet with arbitrary weight kernel size. 
% \todo{Shilin: state that we can apply this to other BNNs as well other than BinaryNet.}

The prototype is implemented by using high-level synthesize tool Xilinx SDx 2018.1. The accelerated functions are written in high-level programming language. The SDx synthesize tool can automatically generate essential AXI bus for memory communication between off-chip memory and FPGA. The SDx tool also synthesizes the marked function into RTL and bitstream. During the inference, the CPU will awake the FPGA acceleration once the hardware function is called. Our design is implemented on the Xilinx Zynq ZCU104 board containing an ARM Cortex-A53 processor with a target clock frequency of 200MHz. As our design is scalable in the number of PEs, we configure the design with 8 PEs in the following experiments. 
%for the testing of the following section.

\begin{table}
  \caption{Summary of the workload}
  \label{tab:workload}
  \begin{tabular}{ccccc}
    \toprule
    layer & \thead{input dim \\$(h,w)$} & \thead{weight dim \\$(r,s,c,k)$} & \thead{weight size \\(Bits)} & \thead{graph partition parameters\\ $(V,K)$} \\
    \midrule 
    conv1 & 32,32& 3,3,128,128& 144K & 64,2 \\
    pool & 32,32 & - & -& - \\
    
    conv2 & 16,16&3,3,128,256& 288K& 64,4 \\
    conv3 & 16,16&3,3,256,256& 576K& 64,4 \\
    pool & 16,16&-& -& - \\
    
    conv4 & 8,8&3,3,256,512& 1.1M& 64,8 \\
    conv5 & 8,8&3,3,512,512& 2.3M& 64,8 \\
    pool & 8,8& - & -& - \\
    
  \bottomrule
\end{tabular}
\end{table}

\subsection{Performance Analysis}
\label{perf}

\subsubsection{Sensitivity to input similarity}
\begin{figure}[bth]
	\centering
	\includegraphics[width=0.5\textwidth]{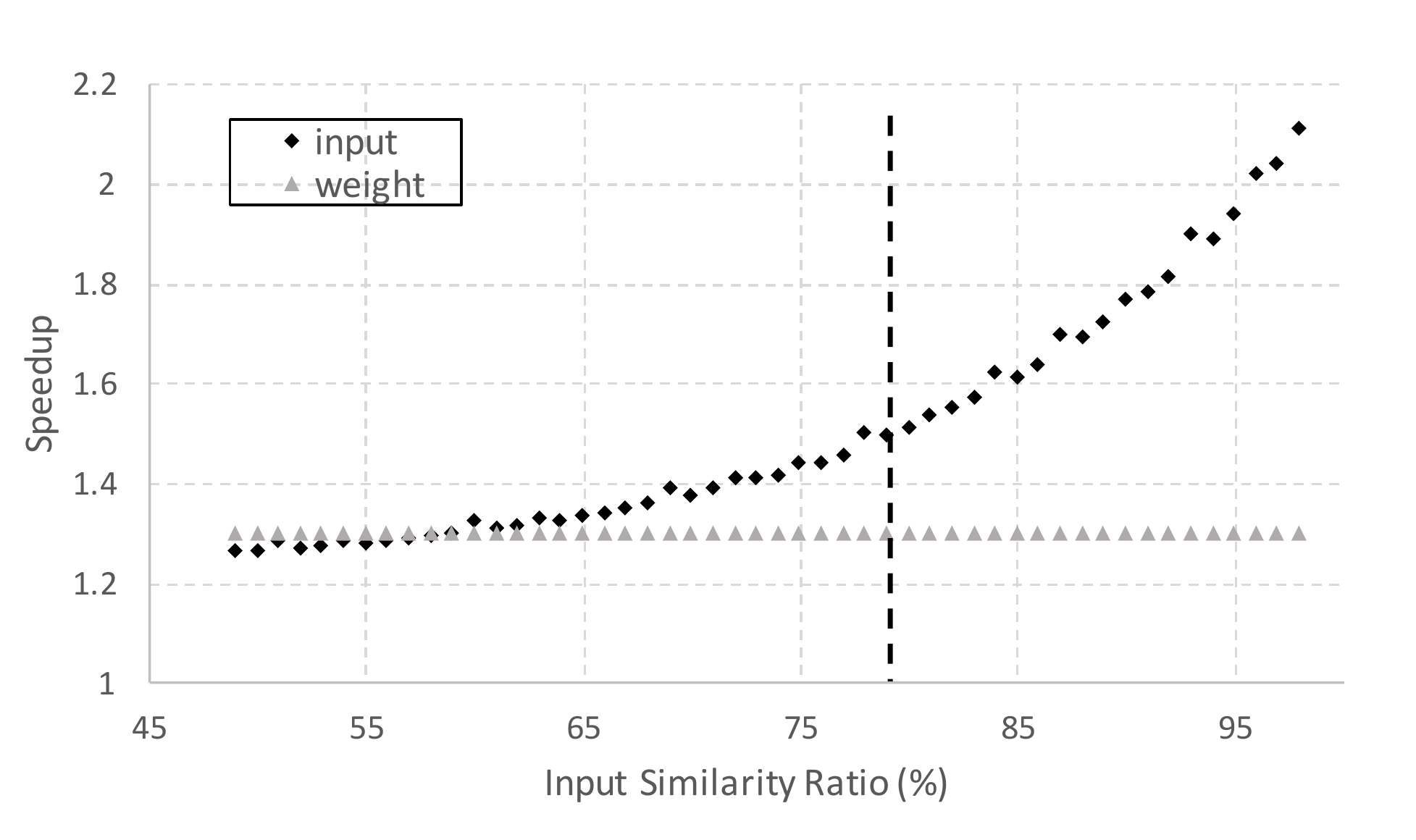}
	\caption{Performance of the accelerators as a function of the input and kernel similarity. The vertical dash line in the graph indicates the average input image similarity among the testing dataset. The baseline is the runtime of input reuse accelerator without using the reuse technique.}
	\label{fig:fig_result_1}
	\vspace{-0.3in}
\end{figure}

To show how we choose between the input and the weight reuse accelerator, we study the influence of input similarity ratio on the performance of the two designs. 
Figure \ref{fig:fig_result_1} shows the speedup of accelerators as a function of the input similarity ratio defined in Section \ref{reorder}. The input reuse accelerator provides a variable speedup which depends on the similarity ratio of the input application. However, the weight reuse accelerator can provide a stable speedup which is based on the similarity ratio between weight kernels after re-ordering. As is shown by the vertical dash line in Figure \ref{fig:fig_result_1}, the speedup at the point of average input similarities of the input applications which is corresponding to the third row of Table ~\ref{tab:sim}, input reuse strategy can provide a better performance compared to weight reuse. Thus, we can conclude that for CIFAR-10 model, input reuse accelerator can provide a better performance and should be used for this BNN architecture. For BNN models with fixed-point input activations, as is shown in the  \ref{tab:sim} where the input is fixed-point value, the similarity ratio for input is low and the decision-making process may prefer weight reuse strategy in such case. The analysis for BNN models which prefer weight reuse is left for future work.

% \todo{Shilin: You may want to echo Figure 1 and emphasize the preference of choosing which reuse strategy. E.g., here if both input and weight is binarized, Figure 4 shows that input reuse is a better strategy. However as we already seen in Figure 1, if we use quantized input to make BNN have better performance (According to BNN research, activation binarization causes severe problems. But weight binarization usually causes little harm), we may prefer weight reuse in such case. blablabla... and this is left for future work.}

\begin{figure}[bth]
	\centering
	\includegraphics[width=0.5\textwidth]{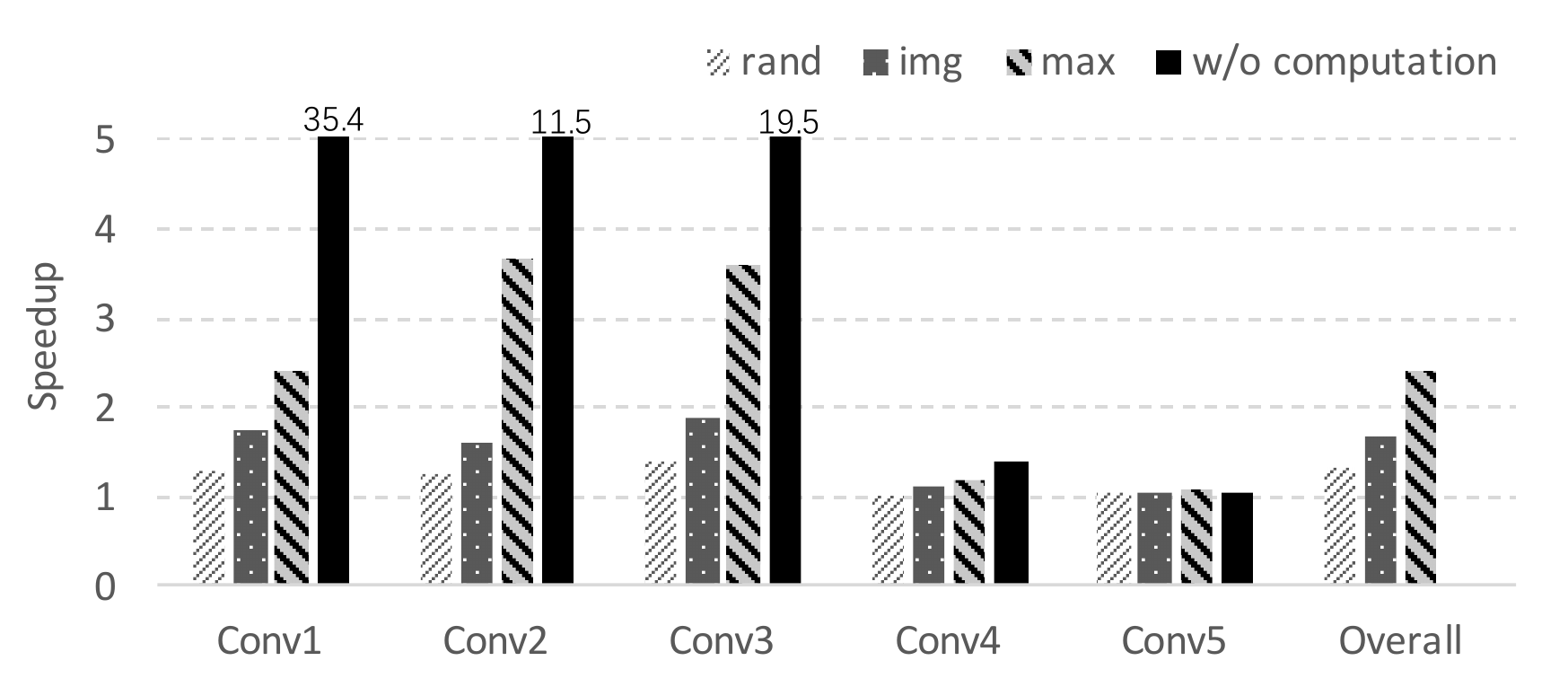}
	\caption{Speedup of the input similarity accelerator across different layers for different types of applications}
	\label{fig:fig_chart1}
	\vspace{-0.1in}
\end{figure}

We also compare the performance of the input accelerator for different types of applications, "rand" indicates the input image is random (-1/+1) series, while "img" is the average performance of the testing images from CIFAR-10 testing dataset, "max" is tested when all the pixels of the input image is in the same color, i.e., all the pixels in the classified input image are the same, "w/o computation" indicates the runtime restricted by off-chip data transfer and CPU control overhead. Figure \ref{fig:fig_chart1} shows the effect of these input applications versus speedup. For conv4 and conv5, the speedup of exploiting the input similarity is small and this is due to that the input activation size is small and weight size is large. We expect to see the bottleneck in these layers is in the off-chip memory bandwidth. 
% To study the effect of the speedup of computation, we removed the bottleneck of DRAM control and memory overhead by preload the data and weight on-chip. The result of   

\begin{figure}[bth]
	\centering
	\includegraphics[width=0.5\textwidth]{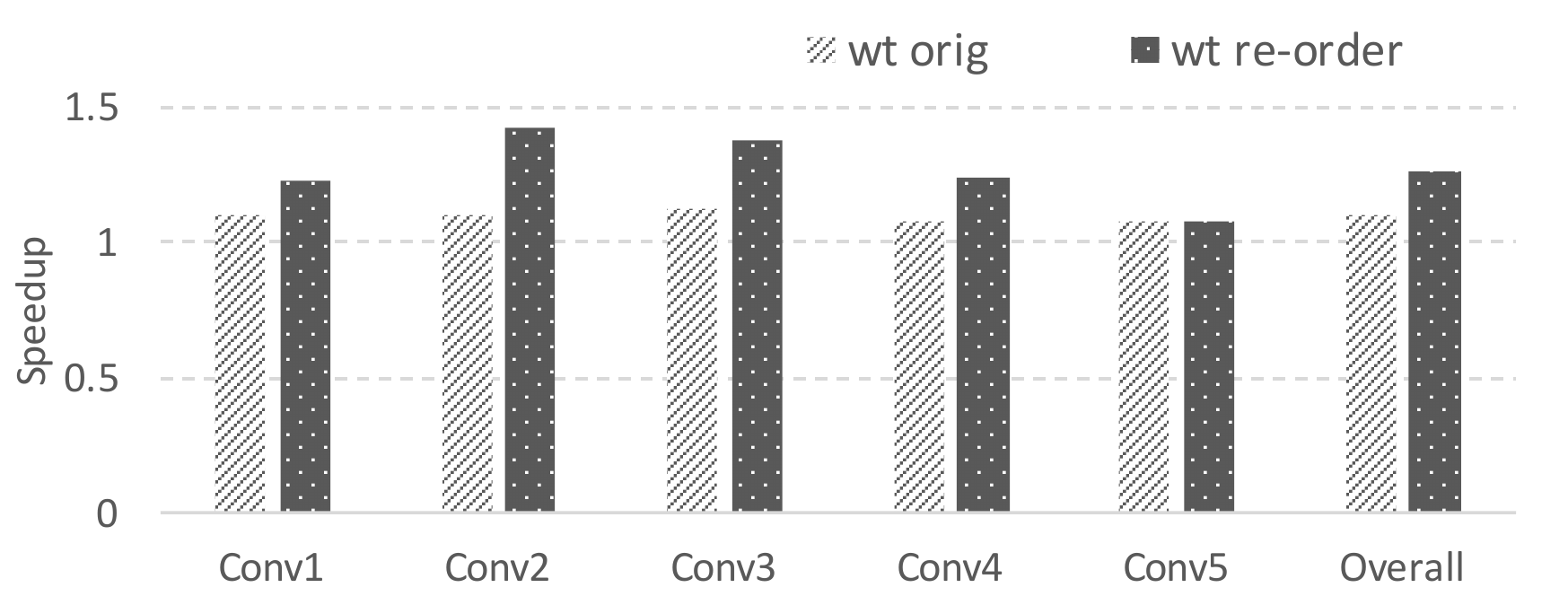}
	\caption{Speedup of using the weight re-order algorithm. The baseline is the weight reuse architecture without using the reuse technique}
	\label{fig:weight}
	\vspace{-0.1in}
\end{figure}

In terms of the speedup by weight reuse, we notice that the similarity ratio of kernel similarity cannot bring too much gain in the speedup. The detail of the speedup of weight re-ordering algorithm is shown in Figure ~\ref{fig:weight}. "wt orig" indicates the performance of original weight order and 'wt re-order' represents the performance of the acceleration applied with off-line reordered algorithm.  By utilizing the re-ordering strategy, the inference can achieve 1.26 $\times$ speedup on average. 

% Figure 7 shows the weight memory access with different input application. 

\begin{figure}[bth]
	\centering
	\includegraphics[width=0.5\textwidth]{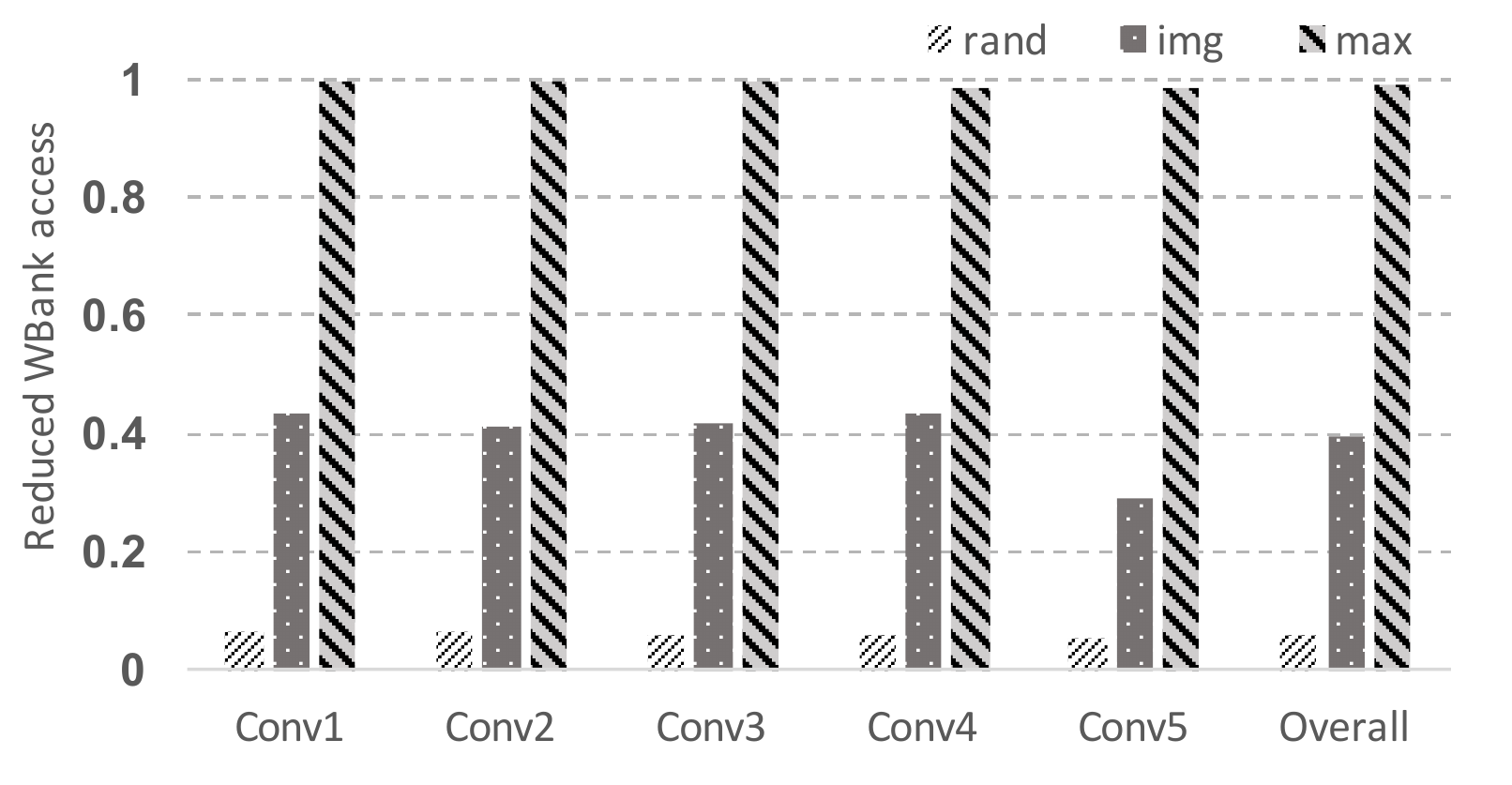}
	\caption{Percentage of reduced weight bank access across different convolution layers }
	\label{fig:fig_w}
	\vspace{-0.1in}
\end{figure}

To show the reduction in weight buffer access and the total number of operations by exploiting the BNN reuse technique, we study the reduction of weight buffer access and bitwise operations as is shown in Figure \ref{fig:fig_w}. The baseline is the weight buffer access and bit operations without using the input reuse technique. As the size of weights on-chip dominates the input activation size, we consider only the weight bank access in this experiment to approximate the total data buffer access in this analysis.

We observe that on average, the input reuse technique saves weight bank access by $40\%$ and bitwise operations by 80\% for the testing images, which can lead to reduction in on-chip power consumption. The accelerator will bypass almost all the weight bank access and bitwise operations if the input application exhibits maximal similarity.

We can conclude that there is a high proportion of on-chip computation redundancy in BNN inference. And this property can be leveraged to reduce the on-chip power consumption and further accelerate the inference of BNN.

% We also analyze the effect of weight reuse accelerators, as shown in Figure ~\ref{fig:weight}. 'wt orig' indicates the original weight order and 'wt re-order' denotes the performance with the off-line reordered\hao{check grammar}. The baseline of the comparison is the weight reuse design without using the reuse technique\hao{what?}. By utilizing the re-ordering strategy, the inference can achieve 1.26 $\times$ speedup on average. The result shows that the kernel similarity among the weight kernels are not high enough to achieve significant improvement in speedup. 

% \todo{Shilin: + by using our architecture. Since there may exists architecture that can gain a lot from weight reuse. Your statement should depend on your architecture, you cannot make general conclusion that which one is always better. Another suggestion: if our architecture is biased on input reuse, we may need to not report and compare to weight reuse since that comparison is unfair if the architecture is biased. ans : The exist strategy for weight reuse is not comparable, they are for fixed-point weight and input.}
\begin{table*}
  \caption{Resource utilization and comparison to prior work.} %\todo{The table is too wide, shall we turn it into a two-column table?}}
  \label{tab:comparison}
  \begin{tabular}{cccccc}
    \toprule
      & FPGA'16~\cite{Qiu} & FPGA'16~\cite{Suda} & FPGA'17\cite{zhao2017accelerating}& Our Design PE8 & Our Design PE16\\
     \hline 
    \midrule
    board & \thead{Zynq \\ XC7Z045} & \thead{Stratix-V \\GSD8} & \thead{Zynq \\ XC7Z020}& \thead{Zynq \\ XCZU7EV} & \thead{Zynq \\ XCZU7EV} \\
    \hline 
    clock(MHz) & 150 & 120 & 147& 200 & 200\\
    \hline    
    precision (bit) & 8-16  & 16 & 1-2& 1-2& 1-2\\
    \hline    
    kLUTs  & 183 & 120* & 46.9& 45 & 72\\
	\hline    
    FF & 128K & no report & 46K & 13K & 19K \\
	\hline    
    DSPs & 780 & 760* & 3 & 5 & 1 \\      
	\hline    
    BRAM & 486 & 1377 & 94 & 112 & 1 \\
	\hline        
	GOPS (conv) & 187.8 & 136.5 & 318.9 & \thead{ Rand 306.6 \\ Img 411.4 \\ Max 539.9 } & \thead{ Rand 713.3 \\ Img 917.7 \\ Max 975.4 } \\
    
% 	\hline        
%     Power & 4.8 & 13 & 1\\    
	\hline            
    \textbf{GOPS/kLUT} & \textbf{1.46} & \textbf{1.14} &  \textbf{6.79}& \textbf{9.14} & \textbf{12.74}\\
% 	\hline            
% 	GOPS/W & 1 & 1 & 1 & 1 & 1\\

\bottomrule
\multicolumn{6}{c}{\footnotesize  * refers to the approximation result from previous paper}
\end{tabular}
\end{table*}

\begin{figure}[bth]
	\centering
	\includegraphics[width=0.5\textwidth]{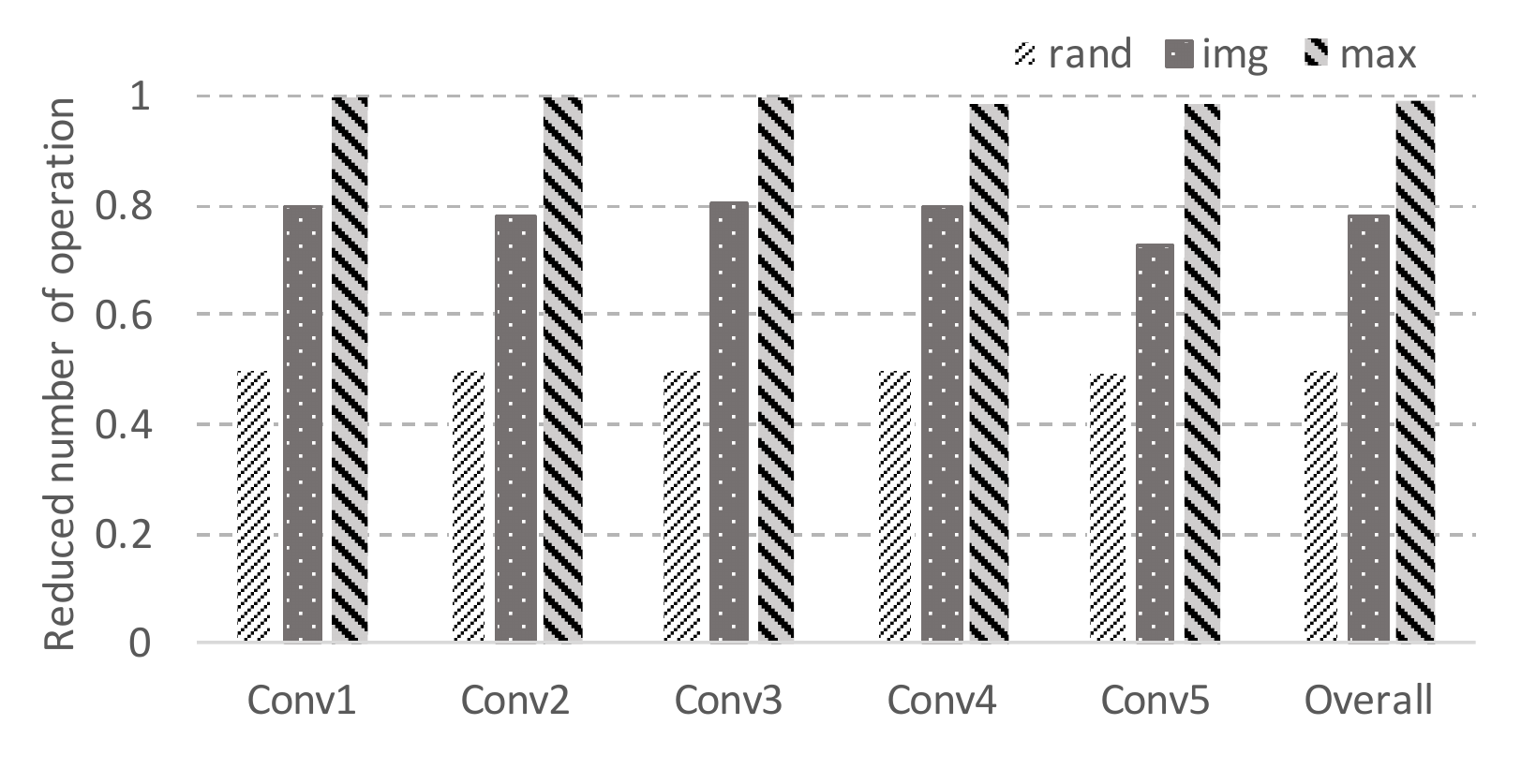}
	\caption{Percentage of reduced number of popcount across different convolution layers}
	\label{fig:fig_pop}
	\vspace{-0.1in}
\end{figure}

\subsection{Power Analysis}

To further analyze the power savings in our accelerator, we study the power consumption which is measured at the socket by using a power monitor. The power consumption of programmable logic is calculated by subtracting the power measured while BNN is running with the power measured at idle stage. 
The measured results are averaged over a period of time while the accelerator is doing required inference.  

\begin{table}
  \caption{Power consumption of the programmable logic}
  \label{tab:power}
  \begin{tabular}{|c|c|c|c|c|c|}
    %\toprule
    \hline
     & w/o reuse &w/o compute & img & max \\
     \hline
    %\midrule
    Power (W)  & 0.36 & 0.21 & 0.30 & 0.23 \\
    \hline
  %\bottomrule
\end{tabular}
\end{table}

Table \ref{tab:power} shows the power consumption of the programmable logic when FPGA is inferencing three different types of applications which are described above in Section ~\ref{perf}. We also added a comparison column which shows the power consumption of data transfer without any on-chip computation. We can conclude that the input reuse strategy on average can reduce the total power consumption by $17\%$ and on-chip power consumption by $54\%$ compared to the baseline accelerator without exploiting the similarity reuse technique.

\subsection{Comparison with Prior Work}

We also compared our results with the state-of-the-art BNN design as shown in \ref{tab:comparison}. As we focused on light-weight architecture for BNN acceleration, we choose a baseline with similar resource consumption for comparison and both of our architectures are for single-bit input and weight. 
% For ~\cite{umuroglu2017finn}, we do not resolve the memory bond by putting the weights on-chip issue and the paper inference a different model. Also, because our input and weight is single-bit, we did not compared with  reference. 
We also put the result with floating point CNN accelerator ~\cite{Qiu} and ~\cite{Suda} here.

Our design is scalable with configurable number of PE. With a large amount of PEs, the on-chip computation will mostly be restricted by the memory bandwidth, the average giga-operations-per-second (GOPS) for testing image becomes closer to the "max" GOPS when we scale the PE size to 16. Our result shows that with PE size of 8 and 16, the design achieves the 9.14 and 12.74GOPS/kLUT, which are 1.34$\times$ and 1.87$\times$ more area-efficient compared to our baseline. The power result is not fair to compare as our FPGA platform of implementation is not the same.

\section{Related works}

\textbf{Binarized Neural Network: }
People have found that it is unnecessary to use floating weights and activations while preserving the accuracy of a neural network. The first approach is to use low-bitwidth fix-point numbers to approximate real values, which is called Quantized Neural Network (QNN) \cite{DBLP:journals/corr/HubaraCSEB16}. However, they cannot fully speed it up because we still need to live with costly operations on multiple bits.

Binarized Neural Network (BNN) was originally proposed in \cite{courbariaux2016binarized} and has received a lot of attention in the research community since bit operations are fast and energy efficient compared to floating-point operations. They have shown the advantage of BNNs in terms of speed, memory usage and power consumption compared with traditional floating number CNN. Many recent works have been proposed to cure BNN's optimization problem during training \cite{xnornet, zhou2016dorefa, Binarynet}. Recently people use ensemble strategy to build a strong binarized network to make BNN both accurate and robust \cite{zhu2018binary}. In this work, we mainly focus on further accelerating BNN inference on FPGA, and our method can be applied to any state-of-the-art BNN architectures.

%\subsection{Hardware Acceleration on Neural Networks}
%\todo{Shilin: this is too fat.}
% \todo{Shilin: Put this section into related works. This is not motivation and background. For background, you need to illustrate how CNN/BNN is implemented on logical arrays, why FPGA is more suitable for neural nets, etc. This part is how previous FPGA work has done on BNN or CNN. THe related work part talks about the ASIC design or other solutions for BNN / CNN inference}
 %-> double check what JZ wrote Ans: This is correct}. 
%Many prior works have been done on FPGA CNN acceleration.
%\hao{move most of the next two paragraphs to the related work section}

\textbf{FPGA acceleration for CNN: } 
FPGA acceleration of CNNs is gaining increasing attention due to the promising performance and cost efficiency.
%Some of the most recent works include 
For instance, Zeng et al.~\cite{Zeng} proposed to employ frequency domain technique to accelerate floating point CNN. 
%A recent study~\cite{7827589} develops a uniform matrix multiplication to accelerate BNNs. 
Escher et al.~\cite{Escher2018} proposed a design that accelerates CNN by optimizing the on-chip storage through identifying optimal batch size for CNNs. Previous work~\cite{7577308} proposed a layer pipelined structure for accelerating large-scale CNN. Ma et al.~\cite{7577356} proposed a compiler design for a scalable RTL to accelerate CNN. Qiu et al.~\cite{Qiu} adopted singular values decomposition to reduce fully-connected layer bandwidth restriction. A recent study~\cite{7827589} develops a uniform matrix multiplication to accelerate CNNs. Zhang et al.~\cite{Zhang:2015:OFA:2684746.2689060} proposed the roofline model which illustrate the computation and memory bond for CNN acceleration and used the model to find the best configuration acceleration.

\textbf{FPGA acceleration for BNN: } 
%For BNN acceleration on FPGA,
Several recent studies explore FPGA acceleration of BNNs.
FINN \cite{umuroglu2017finn} resolved the memory bond issue by storing all the parameters on-chip. Nakahara et al.~\cite{Nakahara} presented a modified BNN version of YOLOv2 to perform real-time localization and classification. ~\cite{Liang:2018:FPB:3198485.3198711} proposed a Resource-Aware Model for optimizing on-chip resource by quantized some part of the input activation. Kim et al.~\cite{kim2017kernel} proposed a kernel decomposition method for BNNs to reduce computation by half. 

% While many previous studies assume that the input are fixed point values, we observe that many of the BNN models are trained with 
% %are assuming 
% single-bit input and weights. 
%%%%JZ: Related work = share the key words of what we proposed = works on either of the following (1) BNN acceleration; (2) input reuse and/or weight reuse. So we don't need to discuss DNN or CNN acceleration.
%There are many 
%Various prior works strive to accelerate BNNs in various platforms.
%FPGA
Li et al.~\cite{Li:2017:EFA} developed an FPGA-based BNN accelerator by leveraging the look-up table (LUT) resources on FPGAs. In order to efficiently implement the normalization and binarization in BNNs with FPGA's LUTs, the design merges the two into a single comparison operation. The study also performs a design space exploration in order to model throughput to improve acceleration performance. Although the design optimizes computation resource utilization by LUT-based computation, the performance is still susceptible to data access bottlenecks~\cite{Zhang:2015:OFA:2684746.2689060}. 

% and are state-of-the-art floating CNN acceleration which are compared in the work. 

% \todo{Also talk about FPGA'16 and FPGA'18. Even though we compared in our results and mentioned in other sections, we need to discuss again here as they are closely related works.} \todo{Instead, our design xxx. As shown in our experiment results, xxx.}
\textbf{ASIC acceleration for BNN: } 
% ASIC
XNORBIN~\cite{DBLP:journals/corr/abs-1803-05849} and Conti et al.~\cite{8412533} implemented Application-specific integrated circuit (ASIC) based BNN accelerators. Their designs adopted loop unrolling and data reuse to exploit the inherent parallelism of BNNs. However, the design simply maps BNN algorithms onto hardware. As a result, their performance improved over traditional neural network accelerations implementations is due to the efficiency of native BNN algorithms. YodaNN ~\cite{7560203} and BRein Memory ~\cite{8226999} are proposed ASIC accelerators for accelerating BNN inference. 
% \todo{Discuss the downside of these works, in comparison with ours.}

\textbf{Computation reuse for CNN: } 
%Near-data processing
%Recent works strive to mitigate the data access bottleneck by emerging in-memory processing techniques. Liu et al.~\cite{Liu:2018:PSA:3195970.3196089} implemented BNN processing in SRAM arrays to exploit the high parallelism in computation. Each 6T bit cell is customized to 8T to support XNOR operations; sense amplifiers are replaced by multi-bit ADCs. Although the design mitigated the data access bottleneck, the customized 8T cells can impose significant scalability and reliability issues with large-scale network models.
%Many works like [][] provided flexible bitwidth for both weight and input to accelerate DNN inference due to the different bitwidth requirements among different CNN applications. [] and [] are work that leverage the sparsity of the DNN. Prime [] provides a solution for DNN acceleration by using resistive RAM. For FPGA CNN acceleration, Escher[1] optimized the batch size of inference to balance the on-chip computation and off-chip memory access. [] and [] provides 
Computation reuse strategies have been proposed for DNN. Marc et al. ~\cite{Marc2018} exploits input similarity between frames to reduce the computation of fixed point DNN inference. The reuse strategy quantizes the input first before checking the value with previous input. Thus, the method will sacrifice a little bit of classification precision. UCNN ~\cite{hegde2018ucnn} quantizes the weight by using TTQ (or INQ)  strategy, so only 3 (or 17) possible values of weight are available in the network. They sort the weights off-line based on weight values to factorize dot product and reduce computation power consumption. 
%YodaNN [54] and BRein [55] propose an ASIC accelerator for accelerating BNN inference. For FPGA BNN accelerator, we are the first work that has exploited the similarity in weight and input activations.
% They sort the weights off-line based on their values to reduce computation and save energy which stores location of input for each weight. 

These prior works on BNN and FPGA, ASIC for accelerating CNNs have enlightened the path of developing high-performance and energy-efficient neural network acceleration. To our knowledge, this is the first paper to exploit similarity in kernels and input activations to effectively accelerate BNNs on FPGA.
% \todo{ Shilin: Topics: 1. Discuss in which environment do we use which reuse strategy, weight or input. 2. Discuss possible combining solution of both input and weight reuse. 3. Discuss limitation of current architecture and say that it can be further optimized. Ans: added the content to the conclusions and future work.}

\section{Discussion}
%\todo{Shilin: Cheng, I will write these and please revise it.}
\textbf{Comparison between input and weight reuse: }
In this paper, we show better performance in speed and energy by using the two types of reuse strategies, i.e., input and weight reuse. From Table ~\ref{tab:sim}, we also notice their differences in different datasets and network models. Generally speaking, the input activation binarization is causing much more harm to performance over weight binarization in BNN, and many high-performance BNNs still prefer using floating or quantized input \cite{zhou2016dorefa}. In such case, weight reuse seems to be a better strategy since it can guarantee a higher degree of similarity. But for BNN where input activation is binarized as well \cite{xnornet}, input reuse is highly preferred for most image classification tasks. We believe more exploration can be made on smartly switching these two strategies and also studying whether the acceleration variance would cause problems in real-time applications or not.

\textbf{Combination of both reuse strategies: }
Another important perspective brought by this paper is to study the mixing architecture which combines these two reuse strategies together in order to gain advantages from both. There may be extra overhead to realize this combination since the hardware architecture could be different. One of the promising next step is to design a more complicated architecture on FPGA that can efficiently accelerate inference by maximizing total similarities. 

\textbf{Improving current inference architecture: } 
%\todo{Shilin: Cheng, please list some limitations of our architecture here and provide some potential ways of improvement.}
The ideal architecture to reduce computation redundancy should decrease the number of $Ops\_per\_image$ in Equation ~\ref{eqfps} without affecting the \textit{Utilization}. Our proof-of-concept architecture is constrained by the off-chip memory bandwidth under some circumstances. Besides, the control overhead for the reduction must also be considered. This results in a small gap between the performance of our accelerator without using the reuse strategy and the-state-of-the-art design.
%as we did not exploit the spatial reduction in $3 \times 3$ weight kernels As is shown in Table, there is a gap exist between the performance of our accelerator without using the reuse strategy and the-state-of-the-art design.
In our future work, we will exploit the similarity  without affecting the utilization of multipliers. It is also possible to combine the input or weight reuse strategy with some previous BNN acceleration techniques ~\cite{umuroglu2017finn}, for example, storing all the weights on-chip for resolving the memory issue. In addition, the design should remain in low control overhead which saves on-chip resource for more computation units.  
% Our architecture for exploiting the input or kernel similarity is mainly for proof-of-concept and the better design will be left for future work. 

% Our envision of the optimal BNN accelerator is to exploit the similarity ratio of removing the redundancy in computation without harming the utilization of multipliers. Also, the design should remain in low control overhead which saves on-chip resource for more computation units. Our architecture for exploiting the input or kernel similarity is mainly for proof-of-concept and the better design will be left for future work.

\section{Conclusions}
In this paper, we propose a new FPGA-based BNN acceleration scheme, which incorporates both algorithm and hardware architecture design principles.
%network inference architecture on FPGA, 
Our design focuses on reducing latency and power consumption of BNNs by exploiting input and kernel similarities. We have shown that BNN inference has the property of high ratio of similarity in both input and kernel weights. The similarity of the input image comes from the spatial continuity between input pixels. Kernel similarity can be enhanced by applying to a proposed reordering algorithm. With different fixed-point representation for BNN input activation, either input or weight exhibits higher similarity ratio which can be exploited to reduce the bit operations and buffer access. By leveraging these two properties of the BNN, we proposed two types of accelerators, which can be applied to different situations. We also summarized the insights generated by comparing these two accelerators to assist strategy selection and combination. Our experiment shows that the power and speed of BNN inference can be largely improved through reducing computation redundancy. We believe this work makes an important step towards deploying neural networks to real-time applications.
\vspace{-0.1in}
% The current design is limited to inference on $32\times32$ or smaller images. A better architecture design that can more efficiently manage the data and weight parameters to exploit input or kernel similarity is yet to be discovered. Future work includes reducing the control overhead and achieve higher parallelization for computation. Also, it is possible to combine the input or weight reuse strategy with some previous BNN acceleration techniques ~\cite{umuroglu2017finn}, for example, storing all the weight parameters on-chip for resolving the memory bond issue. We also need to investigate the way of mixing the two reuse strategies to leverage the advantages from both simultaneously. 
\section*{Acknowledgement}
This work is done during the internship at Iluvatar CoreX. We thank Tien-Pei Chen and Po-Wei Chou for the guidance on FPGA HLS problems. We would also like to thank Wei Shu and Pingping Shao for many helpful discussions. 
\bibliographystyle{ACM-Reference-Format}
\bibliography{acmart}

\end{document}